\documentclass[letterpaper, 10 pt, conference]{ieeeconf}

\IEEEoverridecommandlockouts

\usepackage{etextools}
\usepackage{amssymb}
\usepackage{makeidx}
\usepackage{amsmath}
\usepackage{bbm}
\usepackage{epsfig}
\usepackage{epsf}
\usepackage{psfrag}
\usepackage{verbatim}
\usepackage{color}
\usepackage{multirow}
\usepackage{tabularx}
\usepackage{booktabs}
\usepackage[tight,footnotesize]{subfigure}
\usepackage{array}
\usepackage{soul}
\usepackage{footnote}
\usepackage{cite}
\usepackage{dblfloatfix}

\usepackage{xcolor}
\usepackage{color, colortbl}
\usepackage[colorlinks,bookmarksnumbered,citecolor=orange,urlcolor=orange]{hyperref}
\usepackage{graphicx}
\graphicspath{{Figures}}
\DeclareGraphicsExtensions{.pdf,.png}
\usepackage{bigstrut}
\usepackage[english]{babel}

\usepackage{csquotes}

\usepackage[T1]{fontenc}
\usepackage{algorithm, setspace}
\usepackage{algpseudocode}
\usepackage{url}
\usepackage{multirow}
\usepackage{float}
\usepackage{xcolor}
\usepackage{hyperref}
 \hypersetup{
     colorlinks=true,
     linkcolor=orange,
     filecolor=orange,
     citecolor=orange,      
     urlcolor=orange,
     }

\usepackage{balance}

\newcommand{\p}[1]{\smallskip \noindent \textbf{{#1}.}}
\newcommand{\eq}[1]{Equation~(\ref{eq:#1})}
\newcommand{\fig}[1]{Figure~\ref{fig:#1}}

\title{\LARGE
Communicating Robot Conventions through Shared Autonomy
}

\author{Ananth Jonnavittula and Dylan P. Losey
\thanks{The authors are members of the Collaborative Robotics Lab (\href{https://collab.me.vt.edu/}{Collab}), Dept. of Mechanical Engineering, Virginia Tech, Blacksburg, VA 24061.
\newline
{e-mail: \texttt{\{ananth, losey\}@vt.edu}}}
}

\begin{document}
\maketitle
\begin{abstract}

When humans control robot arms these robots often need to infer the human's desired task. Prior research on assistive teleoperation and shared autonomy explores how robots can determine the desired task based on the human's joystick inputs. In order to perform this inference the robot relies on an internal mapping between joystick inputs and discrete tasks: e.g., pressing the joystick left indicates that the human wants a plate, while pressing the joystick right indicates a cup. This approach works well \textit{after} the human understands how the robot interprets their inputs --- but inexperienced users still have to learn these mappings through trial and error! Here we recognize that the robot's mapping between tasks and inputs is a \textit{convention}. There are multiple, equally efficient conventions that the robot could use: rather than passively waiting for the human, we introduce a shared autonomy approach where the robot \textit{actively reveals} its chosen convention. Across repeated interactions the robot intervenes and exaggerates the arm's motion to demonstrate more efficient inputs while also assisting for the current task. We compare this approach to a state-of-the-art baseline --- where users must identify the convention by themselves --- as well as written instructions. Our user study results indicate that modifying the robot's behavior to reveal its convention outperforms the baselines and reduces the amount of time that humans spend controlling the robot. See videos of our user study here: \url{https://youtu.be/jROTVOp469I}

\end{abstract}

\smallskip





\section{Introduction}

\begin{figure}[t]
	\begin{center}
		\includegraphics[width=0.9\columnwidth]{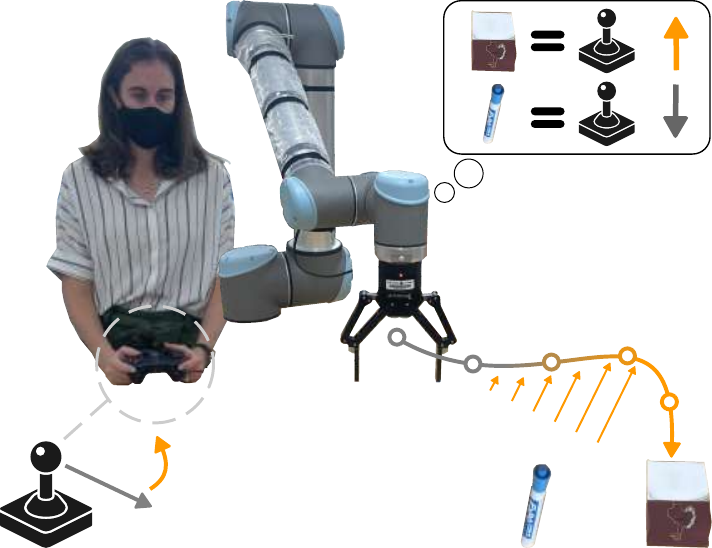}
		\vspace{-0.5em}
		\caption{Human interacting with a joystick to convey their desired task. The human initially presses their joystick straight towards the notepad. But the robot has in mind a convention for interpreting the human's inputs --- \textit{up} for the notepad and \textit{down} for the marker. To communicate this convention, we leverage shared autonomy to modify the robot's trajectory and reveal more informative motions. Humans that gradually adapt their joystick inputs to match these motions will more concisely and accurately convey their intent.} 
		\label{fig:front}
	\end{center}
	\vspace{-2em}
\end{figure}

Imagine teleoperating an assistive robot arm to reach for a notepad in a cluttered environment (like the one in \fig{front}). You interact with a joystick, and the robot leverages these joystick inputs to infer which task you are trying to complete. Intuitively, you might press the joystick \textit{straight} towards your desired object. Although this input makes sense to you, it could confuse the robot arm: there is a marker right next to the notepad, and the robot is not sure which of these two objects you really wanted to reach.

To interpret human inputs and predict their desired task robots use an inference or intent detection algorithm. At its heart, this inference algorithm is based on a robot-assigned \textit{convention}: i.e., a mapping between high-level tasks and low-level inputs. For example, the robot could assume that the human will directly aim their joystick towards their desired goal. We refer to this model as a convention because there are \textit{multiple, equally optimal} mappings from tasks to inputs. Returning to our example, another convention could be moving the joystick \textit{up} to indicate the notepad, and \textit{down} to indicate the marker --- pressing up (or down) is no more challenging than aiming straight for the notepad.

Understanding the robot's convention is key to communicating with the robot. Once the human knows how the robot interprets their inputs, the human can then follow this convention to seamlessly convey their intent. We specifically focus on \textit{shared autonomy} settings --- here the standard convention is for the human to move the robot arm directly towards their task \cite{dragan2013policy, jain2019probabilistic, javdani2018shared, jeon2020shared, jonnavittula2021learning, zurek2021situational, brooks2019balanced}. At first glance, this makes inference challenging when the potential tasks are close together \cite{fontaine2020quality}: a small mistake could point the joystick towards the marker instead of the notepad. But this confusion is easily avoidable once the user understands the robot's convention: the human can reliably convey their task by exploiting the convention and aiming to the right of the notepad, clearly avoiding the marker. This choice of action that unambiguously indicates the human's task is an \textit{exaggeration}.

In this paper we explore how robots can convey their built-in conventions to inexperienced human users. We recognize that --- because the robot knows its own conventions --- it also knows how humans \textit{should} interact with these conventions to seamlessly communicate their task. Specifically in the context of shared autonomy, our insight is that:
\begin{center}\vspace{-0.3em}
\textit{Robots can reveal their conventions by guiding the human's behavior towards more communicative inputs.}\vspace{-0.3em}
\end{center}
Robots that apply our insight leverage shared autonomy to actively demonstrate their conventions to the human. This process is shown in \fig{front}: as the human moves directly towards the notepad, the robot intervenes to (a) help the human complete the task while simultaneously (b) guiding the human along a trajectory that would more clearly indicate the notepad. Our central hypothesis is that the human will \textit{learn} from this guidance: the next time they encounter this scenario, the human should update their joystick inputs to mimic the demonstrated behavior. If successful, our approach reduces the amount of time users spend interacting with the joystick to convey their desired task.

Overall we make the following contributions:

\p{Formalizing Conventions in Shared Autonomy} We formulate the role of conventions in inferring the human's desired task. We then enable robots to exploit these conventions and identify the exaggerated policy that the human \textit{should} follow to indicate their task with fewer joystick inputs.
    
\p{Communicating Conventions over Repeated Interaction} We leverage shared autonomy to guide users towards more communicative policies. Across repeated interactions our approach attempts to infer the human's task and then suggests an improved way to indicate that task. We prove that --- if the human mimics the robot --- our approach is more efficient than letting the human find the convention on their own.

\p{Comparison to Written Instructions}
We test our resulting algorithm in scenarios where the human is reaching for objects on a cluttered table or performing continuous skills. We compare with other teaching modalities, including written, crowd-soured descriptions of the conventions. Our results suggest that using shared autonomy to demonstrate conventions outperforms the alternatives, particularly when the conventions are complex or unintuitive.
\section{Related Work}

\noindent \textbf{Shared Autonomy.} Over 13\% of all American adults have some form of physical disability and need assistance during activities of daily living \cite{taylor2018americans}. Robot arms can help these adults perform everyday tasks without relying on caregivers \cite{argall2018autonomy}. Rather than forcing the human to constantly teleoperate the robot arm, it is often beneficial to automate parts of the task \cite{bhattacharjee2020more, gopinath2016human, park2020active}. Shared autonomy arbitrates between the human operator and autonomous assistance so that both agents control the robot's motion. We specifically focus on shared autonomy paradigms where the robot is given the discrete set of candidate tasks \textit{a priori}: during interaction, the robot tries to infer the human's current task and then takes over to autonomously complete the motion \cite{dragan2013policy, jain2019probabilistic, javdani2018shared, jeon2020shared}. Recent work on shared autonomy has focused on how the robot can gather information and learn new tasks from human interactions \cite{jonnavittula2021learning, zurek2021situational, brooks2019balanced}. By contrast, our work explores how robots can \textit{convey information} through shared autonomy in order to improve the human's interactions.

\p{Conventions} When there are multiple optimal solutions to a multi-agent problem, conventions determine which solution the team follows \cite{boutilier1999sequential, shih2021critical, menell2016cooperative}. As an example, we follow a convention to drive on the right (or the left) side of the road. Within this paper, conventions refer to the mapping between the human's desired task and their joystick inputs. Because the task space is discrete and the input space is continuous, there are an \textit{infinite} number of possible conventions: returning to \fig{front}, the human could press the joystick up, down, left, right, or any other angle to indicate the notepad. But for this convention to function, the human must understand what the robot expects. Recent works have explored how robots can learn human conventions \cite{shih2021critical}, how robots can avoid conventions \cite{hu2020other}, and how robots should respond to humans who know the robot's convention \cite{milli2020literal}. Rather than causing the robot to adapt to the human, we study how the robot can drive \textit{human adaptation} to the robot's convention. 

\p{Algorithmic Teaching} To determine how the robot should teach conventions to the human we utilize recent work on algorithmic teaching (also called machine teaching) \cite{zhu2015machine}. Within contexts where the robot is learning from demonstrations, prior work improves human teaching by providing \textit{verbal or written teaching guidance} --- i.e., heuristic instructions on how to teach \cite{cakmak2012algorithmic, cakmak2014eliciting, sena2020quantifying}. However, within our shared autonomy context we propose to leverage the \textit{robot's motion} to teach conventions. Our approach draws from related research on legible motions for human-robot collaboration: here the robot purposely follows an exaggerated trajectory to make its intent clear to onlooking humans \cite{dragan2013legibility, dragan2015effects}. We take inspiration from legible motions and machine teaching to establish a method by which the robot communicates its convention in shared autonomy settings.

\smallskip

Most related to our research are \cite{nikolaidis2017human} and \cite{bragg2020fake}. Both papers focus on shared autonomy, and use robot interventions to change or improve the human's behavior. Like these papers, we will leverage shared autonomy to convey information from the robot to the human --- but unlike these works, our goal is to teach the human how to teleoperate the robot.

\section{Problem Statement}

Let us return to our motivating example where a human is trying to teleoperate the robot arm to pick up a notepad. When the human starts interacting with the robot they know their task (i.e., the human knows that they want the notepad), but they do not know the robot's convention, and therefore they do not know the most \textit{efficient} way to communicate to the robot that they want the notepad. Conversely, the robot knows its convention --- and the most efficient way for the human to indicate either task. However, the robot does not know whether the human wants the notepad or the marker.

\begin{figure*}[t!]
	\begin{center}
		\includegraphics[width=2\columnwidth]{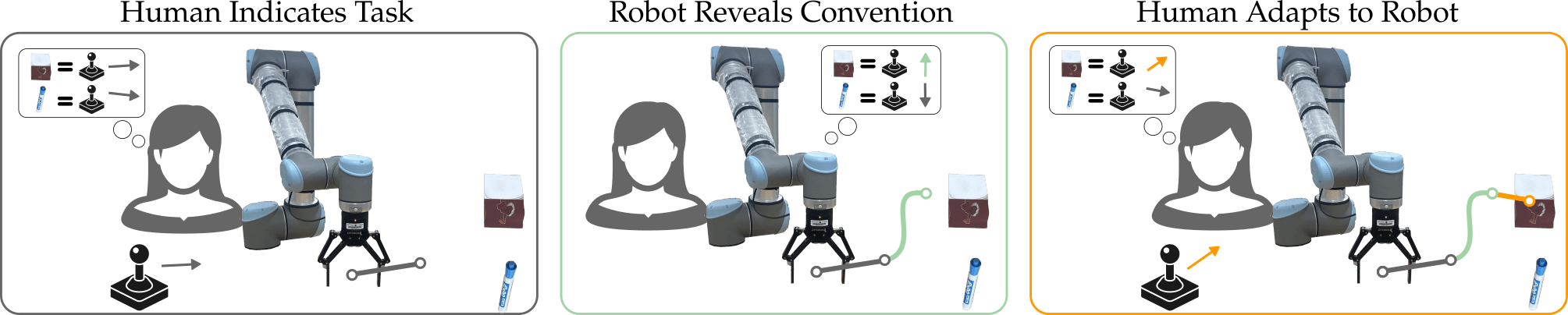}
		\vspace{-0.5em}
		\caption{Our proposed approach for communicating conventions through shared autonomy. (Left) At the start of each interaction the human uses their current understanding of the convention to try and indicate their desired task. (Middle) The robot infers which task is most likely and then intervenes to reveal its convention for that task. Here the robot moves up because pressing up on the joystick would increase its confidence in the notepad. We constrain the robot's motion so that it still assists the human to complete the task. (Right) Our hypothesis is that the human will respond to the robot's intervention by mimicking the robot's behavior. Within this example, the human learns to press their joystick up to indicate that they want to reach the notepad.}
		\label{fig:method}
	\end{center}
    \vspace{-2em}
\end{figure*}

\p{Formalism} Accordingly, we are faced with an \textit{asymmetry of information}. The human has a task (that the robot has to figure out) and the robot selects the convention (which the human cannot observe). We formulate this scenario as a cooperative inverse reinforcement learning (CIRL) game \cite{menell2016cooperative}, i.e., a two-player Markov game where both human and robot receive the same reward: $\mathcal{M} = \langle \mathcal{S}, \{\mathcal{A}_\mathcal{H}, \mathcal{A}_\mathcal{R}\}, T, \{\Theta, R\}, \gamma \rangle$. Here $s \in \mathcal{S} \subseteq \mathbb{R}^n$ is the robot's joint position, $a_\mathcal{H} \in \mathcal{A}_\mathcal{H}\subseteq \mathbb{R}^n$ is the human's commanded velocity\footnote{The human pushes the joystick, and this input is mapped to a joint velocity. Our approach is not tied to any specific mapping: e.g., pressing the joystick right to move the robot's end-effector along the $x$-axis \cite{newman2018harmonic}.}, and $a_\mathcal{R} \in \mathcal{A}_\mathcal{R}\subseteq \mathbb{R}^n$ is the robot's assistance. Within shared autonomy settings the overall action is a combination of the human's commanded action $a_\mathcal{H}$ and the robot's assistance $a_\mathcal{R}$, so that the dynamics $T(s, a_\mathcal{H}, a_\mathcal{R})$ become:
\begin{equation} \label{eq:dynamics}
    s^{t+1} = s^t + \Delta t \cdot f(a_\mathcal{H}^t, a_\mathcal{R}^t)
\end{equation}
One common instance of \eq{dynamics} is linearly blending the human and robot actions \cite{dragan2013policy, jain2019probabilistic, newman2018harmonic}:
\begin{equation}
    f(a_\mathcal{H}^t, a_\mathcal{R}^t) = \beta \cdot a_\mathcal{H}^t + (1 - \beta) \cdot a_\mathcal{R}^t
\end{equation}
where $0 \leq \beta \leq 1$ arbitrates between the human and robot.

Continuing our CIRL formalism, $\Theta$ is the discrete set of candidate tasks that the human might want to complete (i.e., reaching for the notepad, picking up the marker, or opening a drawer), and $\theta \in \Theta$ is the human's current task (which the robot is trying to infer). The reward function $R = \mathcal{S} \times \Theta \rightarrow \mathbb{R}$ depends on the human's current task. To give an example from \fig{front}, we could specify that $R(s, \theta) = 0$ when the robot reaches the notepad and $R(s, \theta) = -1$ at all other states. The scalar $\gamma \in [0,1)$ in $\mathcal{M}$ is a discount factor.

\p{Conventions} Solving the CIRL game produces a pair of human and robot policies $(\pi_{\mathcal{H}}, \pi_{\mathcal{R}})$ that communicate the human's task and maximize long-term reward \cite{menell2016cooperative}. However, choosing this policy pair becomes challenging when there are multiple, equally efficient ways to communicate the human's goal \cite{boutilier1999sequential}. Let $\pi_\mathcal{H}(a_\mathcal{H} | s, \theta)$ be the human's policy: given the robot state $s$ and the human's task $\theta$, this policy determines which joystick input (i.e., which commanded action $a_\mathcal{H}$) the human will provide. We focus on scenarios where there are \textit{a set of equally optimal human policies} $\pi_{\mathcal{H}} \in \Pi_{\mathcal{H}}$. More formally, we consider settings where there are multiple solutions to the CIRL game. The choice of which policy (i.e., which solution) to use determines the team's \textit{convention}.

Consistent with prior work on shared autonomy, here the robot picks the convention \cite{dragan2013policy, jain2019probabilistic, javdani2018shared, jeon2020shared}. In practice, this means that the robot selects some $\pi_{\mathcal{H}} \in \Pi_{\mathcal{H}}$ that it expects the human to follow, and the robot infers the task $\theta$ based on this model. Returning to our motivating example from \fig{front}, the convention could be $\pi_{\mathcal{H}}(a_\mathcal{H} = up \mid s, \theta = notepad) = 1$. We reiterate that there are multiple optimal conventions, i.e., there are multiple policies that convey the human's goal while minimizing the human's effort. For instance, flipping the convention so that down indicates the notepad\footnote{Although our examples involve only a single joystick input, conventions in complex environments may require a sequence of human commands.}: $\pi_{\mathcal{H}}(a_\mathcal{H} = down \mid s, \theta = notepad) = 1$.

\p{Inference} The purpose of establishing a convention is to enable the robot to infer the human's task $\theta$. Recall that the human has a specific task that they want to accomplish, and the robot needs to infer that task. We denote the robot's belief over the discrete set of candidate tasks as:
\begin{equation} \label{eq:belief}
    b^{t+1}(\theta) = P\big(\theta \mid (s^0, a_\mathcal{H}^0), (s^1, a_\mathcal{H}^1), \ldots, (s^t, a_\mathcal{H}^t) \big)
\end{equation}
This belief captures the likelihood of task ${\theta \in \Theta}$ given the history of robot states and human actions. Following \cite{osa2018algorithmic, ziebart2008maximum}, we assume that the human's inputs $a_\mathcal{H}$ are conditionally independent given $s$ and $\theta$. Applying Bayes' rule:
\begin{equation} \label{eq:bayes}
    b^{t+1}(\theta) \propto \pi_{\mathcal{H}}(a_\mathcal{H} \mid s, \theta) \cdot b^t(\theta)
\end{equation}
Hence, the robot's convention $\pi_{\mathcal{H}}$ (i.e., the robot's chosen model of the human's policy) determines how the robot interprets human inputs and infers $\theta$. Within shared autonomy two common instantiations of $\pi_{\mathcal{H}}$ are the Boltzmann rational model \cite{ziebart2008maximum} and the cosine similarity between the human's commanded action and the optimal action for a given task. Both of these conventions expect the human to point their joystick directly towards their target \cite{jain2019probabilistic, javdani2018shared, dragan2013policy, jeon2020shared, zurek2021situational, brooks2019balanced, gopinath2016human}.

\p{Robot} We want to develop an approach that works across arbitrary conventions. Hence, we leave the robot's convention $\pi_{\mathcal{H}}$ as a general human model that maps between high-level tasks and low-level joystick inputs. Recalling that there are multiple solutions $(\pi_{\mathcal{H}}, \pi_{\mathcal{R}})$ to our CIRL game, the robot now executes the policy $\pi_{\mathcal{R}}$ that pairs with convention $\pi_{\mathcal{H}}$. Returning to our running example, let $\pi_{\mathcal{H}}(a_\mathcal{H} = up \mid s, \theta = notepad) = 1$. Accordingly, if the human presses their joystick up, the robot's correct response is to autonomously guide its arm to the notepad. But for the robot to provide the right assistance, it must first understand what the human wants --- and to do this, the robot must teach the human to follow its chosen convention.
\section{Revealing Robot Conventions}

Our proposed approach for revealing the robot's convention is based on \textit{shifting} the human's behavior across repeated interactions (see \fig{method}). The first time that they interact with the robot, the human leverages their \textit{own convention} to communicate their task (e.g., pressing the joystick directly towards the notepad). We want to shift this input over time so that the human gradually understands and effectively leverages the \textit{robot's convention}. In this section we introduce a constrained optimization approach to \textit{\textbf{generate}} actions that reveal the robot's convention. We then explore the conditions the human must satisfy to \textit{\textbf{adopt}} this convention, and prove that demonstrating the convention is more efficient than waiting for the human to learn by themselves.

\subsection{Generating Revealing and Assistive Actions}

To reveal the robot's convention we modify the motion of the robot arm so that --- if the human provides joystick inputs that match the demonstrated motion --- the human will follow the robot's convention. Recall that $b$ is the robot's belief over the discrete set of candidate tasks $\Theta$, and let $\theta^* = \max_{\theta \in \Theta} b^t(\theta)$ be the human's most likely task at the current timestep $t$. Here we optimize for actions that reveal $\theta^*$. Put another way, we seek the commanded human action that will most effectively increase the robot's confidence in $\theta^*$. Within our formalism this action maximizes $b^{t+1}(\theta^*)$, the robot's belief in task $\theta^*$ at the next timestep.

So far we are describing a straightforward optimization. However, this is made more challenging by our shared autonomy setting. On the one hand, the robot should demonstrate informative actions to the human; on the other hand, the robot needs to assist the human and help them to correctly complete their task. We therefore \textit{constrain} the robot's action to ensure that it still assists the human. Our resulting approach for generating revealing and assistive actions is:
\begin{equation}
\begin{aligned}
a_{\mathcal{R}} = \text{arg}\max_{a \in \mathcal{A}_{\mathcal{R}}} & \quad b^{t+1}(\theta^*)\\
\textrm{s.t.} \quad & V_{\theta^*}(s) - Q_{\theta^*}(s, a) \leq \epsilon    \\
\end{aligned}
\end{equation}
Once we substitute in \eq{bayes} and simplify, we reach:
\begin{equation} \label{eq:opt}
\begin{aligned}
a_{\mathcal{R}} = \text{arg}\max_{a \in \mathcal{A}_{\mathcal{R}}} & \quad \frac{\pi_{\mathcal{H}}(a \mid s, \theta^*)}{\sum_{\theta \in \Theta} \pi_{\mathcal{H}}(a \mid s, \theta)}\\
\textrm{s.t.} \quad & V_{\theta^*}(s) - Q_{\theta^*}(s, a) \leq \epsilon    \\
\end{aligned}
\end{equation}
Here $\pi_\mathcal{H}$ is the convention that the robot wants to reveal to the human: the robot leverages this convention to identify actions that maximize the belief in $\theta^*$. We define $Q_{\theta^*}(s,a)$ as the cumulative reward the robot will receive by taking action $a$ in state $s$, and then optimally completing task $\theta^*$ afterwards (with no human assistance) \cite{javdani2018shared}. Finally, $V_{\theta^*}(s) = \max_{a} Q_{\theta^*}(s, a)$ is the maximum expected reward the robot can achieve if it completes task $\theta^*$ autonomously.

When selecting the hyperparameter $\epsilon \geq 0$ in \eq{opt} the designer chooses how much deviation from the optimal policy is allowable. If $\epsilon = 0$, the robot always takes assistive actions (and never reveals information to the human). By contrast, as $\epsilon \rightarrow \infty$ the robot only shows revealing actions, and does not consider whether these actions help the human complete the task. We note that our constrained optimization approach here is similar to \cite{dragan2013generating}: but unlike \cite{dragan2013generating}, we explicitly encode task performance as a constraint.

\p{Algorithm} Our overall approach is displayed in \fig{method} and Algorithm~\ref{alg}. At each timestep the robot infers the human's most likely task (based only on the human's inputs) and then solves for an assistive action $a_\mathcal{R}$ that reveals the robot's convention for that task. Finally, the robot blends the human and assistive actions and transitions to a new state. If the designer wants to make the robot more or less revealing, $\epsilon(t)$ can be changed based on human performance.

\begin{algorithm}[t]
    \setstretch{1.1}
    \caption{Communicating Robot Conventions}
    \label{alg}
    \begin{algorithmic}[1]
    \State \textbf{Input}: Robot's chosen human convention $\pi_{\mathcal{H}}$, discrete set of tasks $\Theta$, and designer-specified hyperparameter $\epsilon$
    \State \textbf{Precompute}: The $Q$-function for each task $\theta \in \Theta$ 
    \While{task not completed}
        \State Observe human command $a_\mathcal{H}^t$ and state $s^t$
        \State $\theta^* \gets \max_{\theta \in \Theta} b^t(\theta)$
        \State $a_\mathcal{R}^t \gets $ solution to \eq{opt}
        \State $s^{t+1} \gets s^t + \Delta t \cdot f(a_\mathcal{H}^t, a_\mathcal{R}^t)$
    \EndWhile
    \end{algorithmic}
\end{algorithm}

\begin{figure*}[t!]
	\begin{center}
		\includegraphics[width=2\columnwidth]{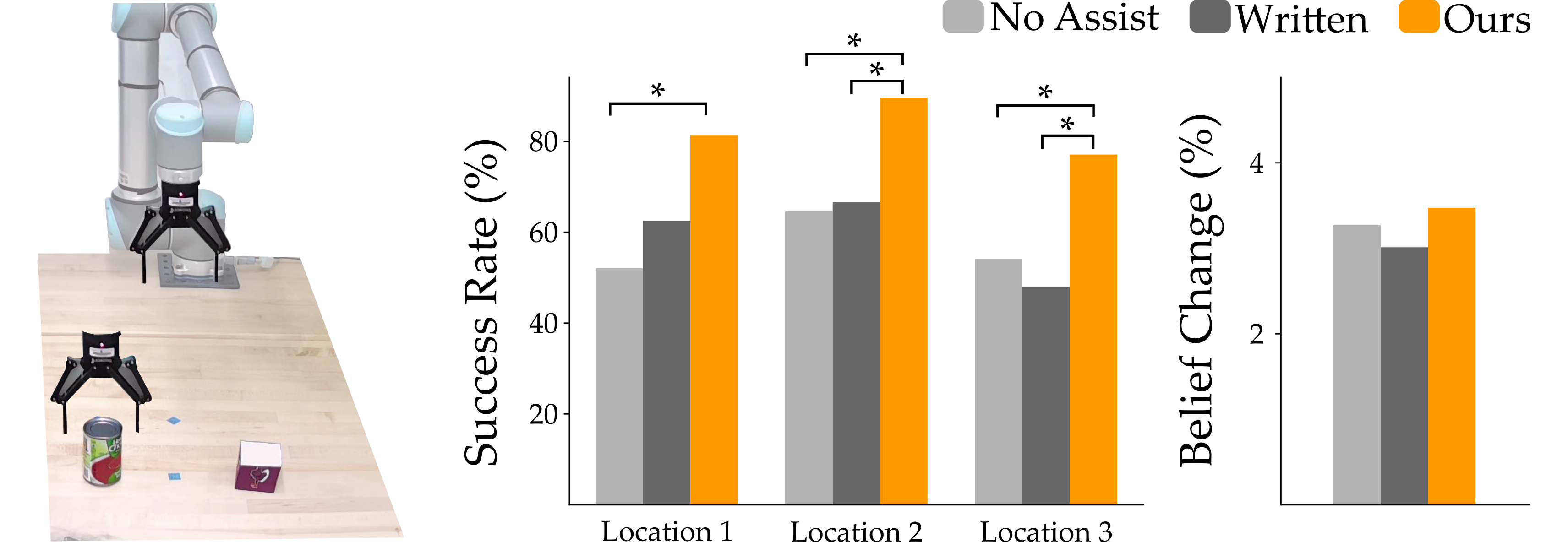}
		\caption{Experimental setup and results for the first and second parts of our user study. (Left) Participants controlled the robot arm to reach for a soup can or notepad. The robot used a Boltzmann rational model to interpret the participants' joystick inputs: under this convention, exaggerated motions more efficiently indicated the desired goal (e.g., pointing the joystick left for the soup can). We explored whether humans adapted to this convention over multiple interactions. (Center) Across three sets of object locations, participants who interacted with \textbf{Ours} were able to convey their desired goal more accurately and concisely. Here $*$ denotes statistical significance $(p < .05)$. (Right) In the second part of our user study we measured how teleoperation behavior changed before and after being exposed to \textbf{No Assist}, \textbf{Written}, or \textbf{Ours}. Participants did improve but the change was minimal. Because we collected the participants' updated behavior in a setting where the robot did not provide assistance, it is possible that users did not feel the need to exaggerate.}
		\label{fig:user1}
	\end{center}
    \vspace{-2em}
\end{figure*}

\subsection{Driving Adaptation to Robot Conventions}

Our approach reveals the robot's conventions to the human --- but is this any more efficient than letting the human find these conventions for themselves? To answer this question we take the human's perspective, and write the setting as a \textit{multi-arm bandit}. The human is interacting with a joystick, and can press this joystick in $N$ different directions: these $N$ discrete inputs become the arms of our bandit. If the human matches the robot's convention and pulls the correct arm (i.e., pushing \textit{up} to indicate the notepad), the human is rewarded by the robot performing the task autonomously. Otherwise, the human has to continually intervene and correct the robot's motion, resulting in more human effort. Define $REG(k)$ as the number of \textit{incorrect} joystick inputs up to interaction $k$.

\p{Without Revealing Actions} If the robot does not actively reveal its convention then the human must explore the space of joystick inputs to find the most effective actions. No matter which policy the human uses to explore these inputs, prior work on multi-arm bandits \cite{slivkins2019introduction, mannor2004sample} has shown that --- in expectation --- the number of incorrect joystick inputs is at least logarithmic in time: $\mathbb{E}[REG(k)] \geq \Omega(\log{k})$.

\p{With Revealing Actions} Our approach to revealing conventions has the potential to lower this bound. But to be effective, the human must actively \textit{explore} different joystick inputs and \textit{learn} from the robot's response. Specifically, we assume that (a) the initial probability of each joystick input is nonzero and (b) the human update their inputs to match the robot's motion. Given these assumptions, the number of incorrect joystick inputs is constant in time: $\mathbb{E}[REG(k)] = C$. This result follows from Proposition~1 in \cite{chan2019assistive} where the roles of the human and robot are reversed: the robot reveals the informative actions for a given task $\theta$ after a finite number of interactions, and thereafter the human mimics the robot's demonstrated convention for that task.

\section{User Study}

Our analysis suggests that demonstrating the robot's convention will help users find and follow that convention more rapidly. But now we need to show that our approach works in practice. Accordingly, we conducted an in-person user study where participants teleoperated a 6-DoF robot arm (Universal Robots UR10) to reach for objects and perform skills.  The study was divided into \textit{three parts} to explore how conventions impact human effort. 



\p{Independent Variables} Over the course of the study participants learned about the robot's conventions: to prevent this from affecting our results, we used a \textit{between subjects} design. Each participant only interacted with one of the following methods: \textbf{No Assist}, \textbf{Written}, and \textbf{Ours} (Algorithm \ref{alg}). 

In \textbf{No Assist} the robot used an existing shared autonomy approach to identify the human's task \cite{dragan2013policy}\footnote{We note that this baseline is interchangeable with other shared autonomy approaches that infer the human's intent.}. This robot never attempted to show its convention to the human: participants using \textbf{No Assist} had to learn the convention through trial and error based on whether the robot assisted for their desired task. In \textbf{Written} the robot also assisted the human without any exaggerations. However, here users were given written, crowd-sourced descriptions of the robot's conventions. These descriptions were obtained from $15$ Amazon Mechanical Turk workers with over a $99\%$ HIT approval rating. As an example, one description told participants to point their joystick ``left and down'' to indicate the notepad. Finally, in  \textbf{Ours} the robot revealed its convention by guiding the human towards more informative inputs.

\p{Experimental Setup} Our user study was divided into the three parts that are described below. Since we followed a between subjects design, each participant completed every part with only one method.

In the first part users were tasked with learning the robot's convention while reaching for either a soup can or a notepad (see \fig{user1}). The robot followed a Boltzmann rational convention \cite{dragan2013policy, jain2019probabilistic, javdani2018shared}. The hyperparameter $\epsilon(t)$ was set to $0.04$ at the start of the task and $\epsilon(t) \rightarrow 0$ as the robot got closer to the goal. Although users could indicate their task by pointing the joystick directly towards their target, exaggerated inputs conveyed the human's task more efficiently (i.e., the human needed fewer joystick inputs to indicate their task if they exaggerated). We tested three object locations: initially the soup can and notepad were located far apart, and then were gradually moved closer together. Participants interacted with each location three times while using their joystick to try and convey the intended goal.

The first part of our user study focused on adapting to a robot convention: in the second part, we tested whether humans would generalize that convention to new scenarios. The soup can and notepad were placed in a new, previously unseen location, and participants teleoperated the robot without any assistance. We measured how the human's teleoperation behavior changed \textit{before} and \textit{after} the first part of our user study. If participants understood the convention that they had experienced in part one, we expected them to provide more informative teleoperation inputs in part two. Regardless of the method used by the participants, the robot only observed the human's input actions and did not apply its policy blending algorithm $f(a_\mathcal{H}, a_\mathcal{R})$ to assist.

Finally, in the third part of our user study we removed the Boltzmann rational model and introduced a less intuitive convention. The robot was able to perform a continuous skill (open a drawer) or reach for an object (soup can). To indicate the drawer, the human needed to input small right and left motions; to indicate the soup can, the human provided larger right and left motions. We tested this unintuitive convention to explore how each method performs when the mapping from joysticks to tasks is more complex. As in part one, participants interacted with the robot three times.

\p{Dependent Measures} For the first part of the user study we calculated the percentage of users who successfully followed the robot's convention (\textit{Success Rate}). To be successful, participants had to (a) complete the intended task and (b) provide fewer joystick inputs than the average across all users. For the second task we calculated how the robot's belief in the human's task changed before and after part one (\textit{Belief Change}). We found the robot's average confidence in the human's true task during their initial interactions, and then subtracted this from the robot's average confidence during the most recent interactions. Hence, this metric captures whether there was a change in how participants teleoperated the robot after experiencing the robot's convention. Finally, in the third part of the user study we again leveraged \textit{Success Rate} to calculated how many times the robot correctly and efficiently inferred the intended task.  

\begin{figure}[t]
	\begin{center}
		\includegraphics[width=1.0\columnwidth]{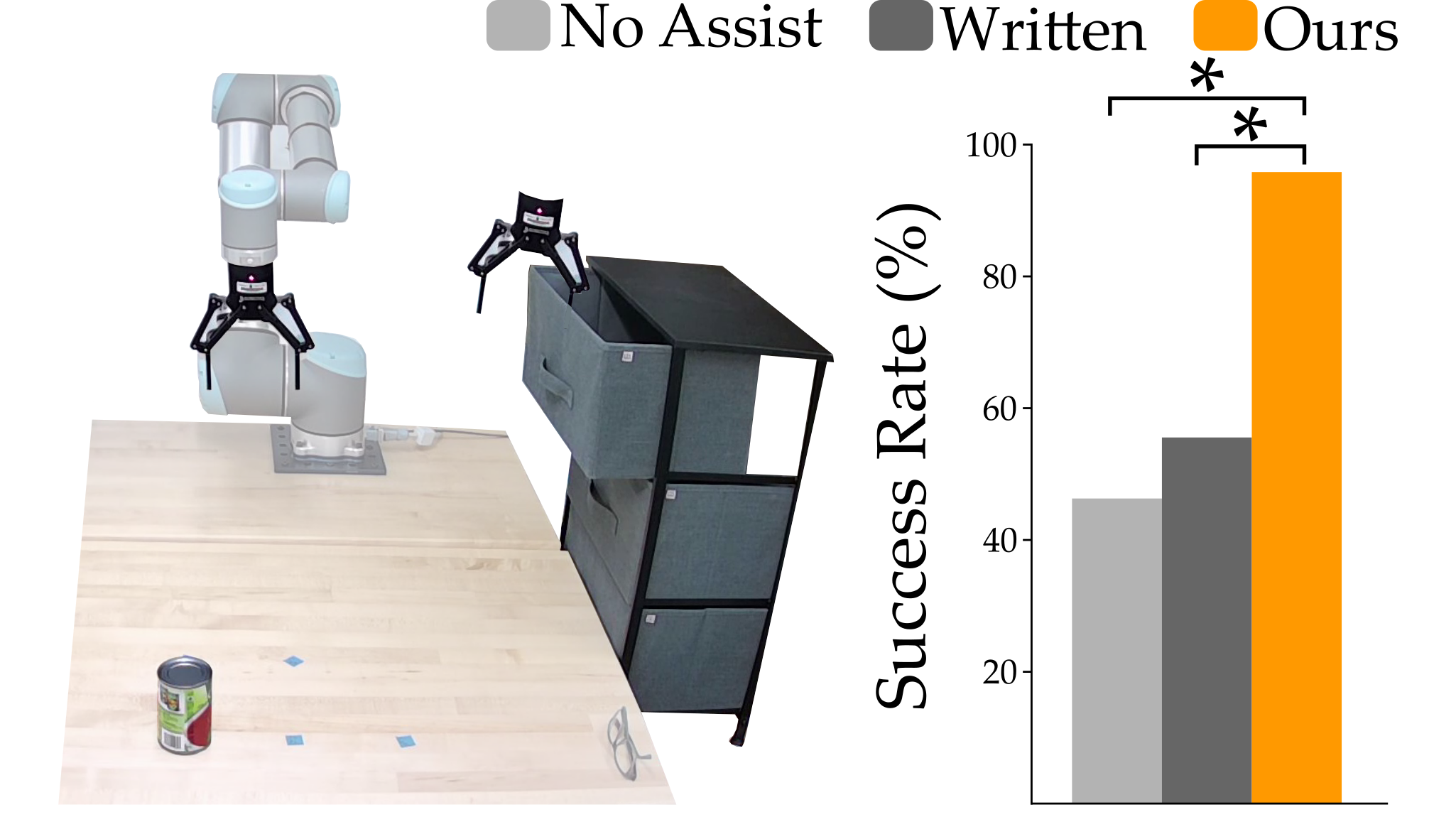}
		\vspace{-0.5em}
		\caption{Experimental setup and results from the third part of the user study. (Left) Here the robot had a complex, sinusoidal convention, and participants needed to match this convention to get the robot to open the drawer or reach the soup can. (Right) We found that participants who \textit{see} the convention (with \textbf{Ours}) outperform participants who \textit{read} the convention (with \textbf{Written}). Here $*$ denotes statistical significance ($p < .05$).} 
		\label{fig:user2}
	\end{center}
	\vspace{-2em}
\end{figure}

\p{Participants and Procedure}
A total of $27$ members of the Virginia Tech community participated in our user study ($8$ female, average age $25 \pm 4$ years). All participants provided informed written consent prior to the experiment. 

\p{Hypotheses}
We tested three hypotheses:
\begin{itemize}
    \item[] \textbf{H1.} \textit{Robots will reveal their conventions by guiding humans towards more communicative inputs.}
    \item[] \textbf{H2.} \textit{Over repeated interactions, humans will generalize these conventions to new scenarios.}
    \item[] \textbf{H3.} \textit{Revealing conventions through robot motion will outperform written descriptions.}
\end{itemize}

\p{Results} The results from the first and second part of our user study are shown in \fig{user1}, and the results from the third part are shown in \fig{user2}.

In the first part of our user study we measured \textit{Success Rate} to see whether participants would adapt to the robot's conventions. Across all $27$ participants, users who interacted with \textbf{Ours} were most likely to follow the robot's convention. Post hoc comparisons between \textbf{Ours} and \textbf{No Assist} were statistically significant for all three object locations ($p < .05$). When comparing against \textbf{Written}, we found that \textit{Success Rate} was significantly higher for locations 2 and 3 ($p < .05$). These results support \textbf{H1}, and suggest that our proposed approach encouraged the participants to exaggerate their joystick inputs and concisely convey their desired task. 

The results from the second part of our user study were not clear-cut. Although users did improve after working with the robot in all conditions, the gains in \textit{Belief Change} were minimal (between $2\%$ and $4\%$). This suggests that participants did not internalize the robot's conventions or transfer those conventions to new scenarios. One possible explanation is that --- because the robot was directly following the human's commanded actions --- participants saw no need to exaggerate and indicate their task. Our results from the second part of the user study do not support \textbf{H2}.

In the third part of the user study we again measured \textit{Success Rate}, but now with the robot following a complex and unintuitive convention. As shown in \fig{user2}, participants who interacted with \textbf{Ours} were best able to match this convention: post hoc analysis confirms that the differences are statistically significant between \textbf{Ours} and \textbf{No Assist} ($p < .05$), and between \textbf{Ours} and \textbf{Written} ($p < .05$). Users who interacted with either \textbf{No Assist} or \textbf{Written} had difficulty learning the robot's unintuitive convention, perhaps because it was easier to \textit{show} than to \textit{tell}.

Looking at each part of the user study, we find support for \textbf{H1} and \textbf{H3}. Robots that leveraged our approach from Algorithm~\ref{alg} were not only able to communicate their convention to the human, but they also communicated this convention more effectively than written instructions. On the other hand, our results do not support \textbf{H2}. None of the methods caused participants to transfer their learned conventions to a new scenario; however, we recognize that this may have been because the robot was not actively inferring their task or assisting the human in this scenario.

\p{Discussion} Our results suggest that users who interact with \textbf{Ours} can be divided into two distinct groups. The majority of participants ($8$ out of the $9$ users) adapted to the robot's convention immediately --- these participants only needed one or two interactions with our approach to understand the mapping from joystick to task. At the other end of the spectrum we also had one participant who \textit{never} adapted to the robot's convention ($1$ of the $9$ users). This participant was often confused by \textbf{Ours}, especially when the robot intervened to modify their motion. Instead of mimicking the robot's behavior, the user applied the opposite input in order to cancel out any autonomous guidance and move directly to the goal. Moving forward, we believe that we can reach this second group by combining both \textbf{Written} and \textbf{Ours}. We hypothesize that the written instructions will provide the context these users need to understand why the robot is altering their motion.
\section{Conclusion}

When humans teleoperate robots there are many conventions the robot can leverage to map joystick inputs to discrete tasks. We have enabled robots to actively communicate their chosen convention. Specifically, we leveraged constrained optimization to identify robot actions that \textit{reveal} how humans should convey the current task while simultaneously \textit{assisting} for that task. Our user study results suggest that this shared autonomy approach reduces the number of joystick inputs that humans need to indicate their task.


\newpage
\balance
\bibliographystyle{IEEEtran}
\bibliography{IEEEabrv,bibtex}

\end{document}